\documentclass[conference]{IEEEtran}
\IEEEoverridecommandlockouts

\usepackage{hyperref}
\usepackage{graphicx} 
\usepackage{caption}
\usepackage{subcaption}
\usepackage{algorithm}
\usepackage{algpseudocode}
\usepackage{booktabs} 
\usepackage{cleveref}
\usepackage[normalem]{ulem}
\usepackage{xcolor}
\usepackage{gensymb}

\def\BibTeX{{\rm B\kern-.05em{\sc i\kern-.025em b}\kern-.08em
    T\kern-.1667em\lower.7ex\hbox{E}\kern-.125emX}}
    \makeatletter
\newcommand{\linebreakand}{%
  \end{@IEEEauthorhalign}
  \hfill\mbox{}\par
  \mbox{}\hfill\begin{@IEEEauthorhalign}
}
\makeatother
\begin{document}

\title{Automated Road Extraction and Centreline Fitting in LiDAR Point Clouds\\
\thanks{979-8-3503-7903-7/24/\$31.00 ©2024 IEEE}
}

\author{
    \IEEEauthorblockN{Xinyu Wang}
    \IEEEauthorblockA{
        \textit{The University of Western Australia} \\
        Perth, Australia \\
        23631069@student.uwa.edu.au
    }
    \and
    \IEEEauthorblockN{Muhammad Ibrahim}
    \IEEEauthorblockA{
        \textit{Department of Primary Industries and Regional Development} \\
        Perth, Australia \\
        muhammad.ibrahim@dpird.wa.gov.au
    }
    \linebreakand 
    \IEEEauthorblockN{Atif Mansoor}
    \IEEEauthorblockA{
        \textit{The University of Western Australia} \\
        Perth, Australia \\
        atif.mansoor@uwa.edu.au
    }
    \and
    \IEEEauthorblockN{Hasnein Tareque}
    \IEEEauthorblockA{
        \textit{Department of Primary Industries and Regional Development} \\
        Perth, Australia \\
        hasnein.tareque@dpird.wa.gov.au
    }
    \linebreakand 
    \IEEEauthorblockN{Ajmal Mian}
    \IEEEauthorblockA{
        \textit{The University of Western Australia} \\
        Perth, Australia \\
        ajmal.mian@uwa.edu.au
    }
}

\maketitle



    



\begin{abstract}
Road information extraction from 3D point clouds is useful for urban planning and traffic management. Existing methods often rely on local features and the refraction angle of lasers from kerbs, which makes them sensitive to variable kerb designs and issues in high-density areas due to data homogeneity. We propose an approach for extracting road points and fitting centrelines using a top-down view of LiDAR based ground-collected point clouds. This prospective view reduces reliance on specific kerb design and results in better road extraction. We first perform statistical outlier removal and density-based clustering to reduce noise from 3D point cloud data. Next, we perform ground point filtering using a grid-based segmentation method that adapts to diverse road scenarios and terrain characteristics. The filtered points are then projected onto a 2D plane, and the road is extracted by a skeletonisation algorithm. The skeleton is back-projected onto the 3D point cloud with calculated normals, which guide a region growing algorithm to find nearby road points. The extracted road points are then smoothed with the Savitzky-Golay filter to produce the final centreline. Our initial approach without post-processing of road skeleton achieved 67\% in IoU by testing on the Perth CBD dataset with different road types. Incorporating the post-processing of the road skeleton improved the extraction of road points around the smoothed skeleton. The refined approach achieved a higher IoU value of 73\% and with 23\% reduction in the processing time. Our approach offers a generalised and computationally efficient solution that combines 3D and 2D processing techniques, laying the groundwork for future road reconstruction and 3D-to-2D point cloud alignment.


\end{abstract}


\section {Introduction}
With the rapid increase in urbanization, there is a dire need for improved urban planning and traffic management. Accurate road information is critical for enhancing the navigation accuracy of autonomous vehicles, optimizing urban resource allocation, and improving emergency response systems \cite{Kootsookos, Bisheng}. Efficient road extraction is also essential for creating up-to-date road structures, monitoring road conditions, and enabling autonomous vehicles to navigate safely.

3D point clouds acquired with LiDAR sensors are commonly used for road mapping in cities \cite{BOYKO2011S2} as well as buildings \cite{Kaminsky}. Surveyors no longer need to rely on handheld instruments for point-to-point measurement, leading to reduced work hours and improved safety \cite{rs10101531}. Extracting roads from images has limitations, as images do not capture 3D structure and are sensitive to environmental factors like lighting and weather changes \cite{9339878, 7287763, UnLoc}, making multiview 3D structure estimation challenging. In contrast, LiDAR sensors offer unique advantages by capturing spatial structures in 3D \cite{s24020503} through active sensing, generating 3D point clouds with rich structural details and fewer external interferences compared to 2D image-based techniques \cite{WeixingWang2016Aror}. Active sensing also makes LiDAR scanning robust to extreme changes in illumination, significantly improving the accuracy and reliability of road point extraction and centerline fitting with LiDAR-based data.

\textcolor{black}{Compared to point clouds collected by airborne LiDAR sensors \cite{article-2, 10092204}, ground-based 3D point cloud data is more suitable for urban areas. Firstly, ground-based scanning provides a more dense coverage of the locality. Secondly, ground-based scanning avoids occlusions caused by tree canopies, bridges and other similar structures. Finally, flying drones in densely populated urban areas is generally prohibited. Keeping in view these reasons, our approach focuses on 3D point clouds collected from ground based sensors.}


\textcolor{black}{Ground-based 3D point cloud data used for road extraction can be divided into two main types.} The first consists of independent frames of point clouds acquired from sensors mounted on vehicles \cite{Wen2008/12}. These are appropriate for real-time navigation and swift environmental analysis, focusing on rapid processing and accurate road information extraction to support navigation in dynamic environments \cite{YANG201380, 5940447}. The second type is large-scale point cloud, which offers a broad coverage and is mainly used in urban planning and management \cite{rs10101531}. This type of point cloud data is usually a merged version of the first type. Such applications typically involve 3D city modelling and extensive terrain mapping \cite{MRENet, 7729407}. Both present unique attributes regarding data volume, coverage, the required processing techniques and processing time \cite{s24020503, 9339878}. 


Many researchers have focused on road extraction from 3D point clouds in urban environments. Suleymanoglu et al. \cite{s24020503} used an activity-driven method that leverages vehicle trajectory data to distinguish road and non-road regions, such as by analyzing GPS trajectories to determine road areas. However, vehicle trajectory data may not be available in some cases. Other research approaches implement feature-driven methods that detect lane lines and road boundaries by extracting distinctive features like curb height, point cloud density projection, and road flatness \cite{9339878, YANG201380, MRENet, inproceedings_1}. Additional approaches extract road surfaces by analyzing point cloud density, elevation differences, and slope changes \cite{BOYKO2011S2, article-2}. Some methods focus on elevation discontinuities at road boundaries to extract curb points \cite{7938403, 7995848, article-1, Qingquan}, but road discontinuity and boundary features can be unreliable due to structural variations. To overcome these limitations, in this paper, we propose an approach for road extraction and centerline fitting from top-down 3D point clouds. Our method combines statistical outlier removal, grid-based segmentation, and 2D skeletonization to achieve promising results without relying on vehicle trajectory data or roadside/curb detection.

The rest of the paper is organised as follows. Section \ref{Related Works} surveys existing road extraction methods. The proposed approach is explained in Section \ref{Methodology}. Experiments and results are discussed  in Section \ref{Experiment}. Section \ref{conclusion} discusses the conclusion and future directions.


\section {Related Work} \label{Related Works}
  

Feature-based methods have been extensively explored in the field of 3D road extraction. Lu et al. proposed combination of machine learning with clustering techniques, such as Density-based clustering algorithm (DBSCAN)~\cite{schubert2017dbscan} and Random sample consensus (RANSAC)~\cite{schnabel2007efficient}, to handle complex road scenarios, including straight segments, turns, and roundabouts \cite{s24020503}. This method has been evaluated on data acquired from Mobile Laser Scanning (MLS) and Mobile Mapping System (MMS). Yang et al. proposed point cloud segmentation and model-driven extraction, where the point cloud was segmented into patches, and a moving window operation was employed to filter out non-ground points \cite{YANG201380}. A model-driven strategy is then used to extract curb points, followed by a refinement and optimisation process to generate the filtered road surface points.  Additionally, an approach that utilises SuperVoxel clustering has been proposed by Mi et al.  for automatic extraction and vectorisation of 3D road boundaries from MLS point clouds \cite{9339878}. This method employs SuperVoxel clustering to extract candidate curb points and applies contracted distance clustering to group boundary segments. The final vectorised road boundaries are generated through an automated extraction pipeline structure including steps such as fitting, tracking, and final completion.

Deep learning methods are actively researched for road extraction and centerline fitting \cite{Balado}. Ma et al. proposed a multi-scale point-based convolutional neural network (CNN) for 3D object segmentation from large-scale LiDAR point clouds \cite{8943956}. The approach employed improved point convolution, a U-shaped downsampling - upsampling structure, and dynamic graph convolution modules. Another deep learning-based approach proposed by Soilán et al. utilises 3D point clouds acquired from low-cost mobile laser scanning systems to parameterise road horizontal alignment \cite{article-4}. The approach used deep learning for semantic segmentation of point clouds and extracted road centrelines with classification. This method also involves parameterising the geometric information of each curve to generate road alignment data.

Hui et al. proposed a road centerline extraction approach that employs skewness balancing, rotating neighborhoods, hierarchical fusion, and optimization algorithms \cite{article-2}. Boyko et al. introduced a method that divides the road network into independent patches, allowing for efficient processing of large datasets \cite{BOYKO2011S2}. Their method uses heuristic techniques such as curb detection to extract road boundaries. Despite these advancements, road extraction from point clouds remains challenging due to varying curb designs \cite{doi:10.1080/19479832.2016.1188860, 10087468} and homogeneity issues in dense areas \cite{10.1145/1653771.1653851}. Changes in road structures and curb designs further complicate road extraction \cite{7287763}.

Feature-based strategies, merging machine learning with clustering \cite{s24020503, MRENet}, and model-driven approaches \cite{YANG201380} have proven effective in complex environments. Deep learning methods, such as point-based CNNs \cite{8943956} and semantic segmentation \cite{article-4}, have demonstrated promising results in road data extraction. Algorithms like Skewness balancing and patch-based processing \cite{article-2, BOYKO2011S2} enhance road extraction efficiency and accuracy in dense point clouds.  Road extraction from point clouds is still a challenging problem due to variations in curb design \cite{doi:10.1080/19479832.2016.1188860}, lack of data uniformity in dense urban areas \cite{10.1145/1653771.1653851}, and the dynamic nature of urban environments \cite{7287763}. Our porposed method overcomes these challenges.


\begin{figure*}[ht]
    \centering
    \includegraphics[width=1\linewidth]{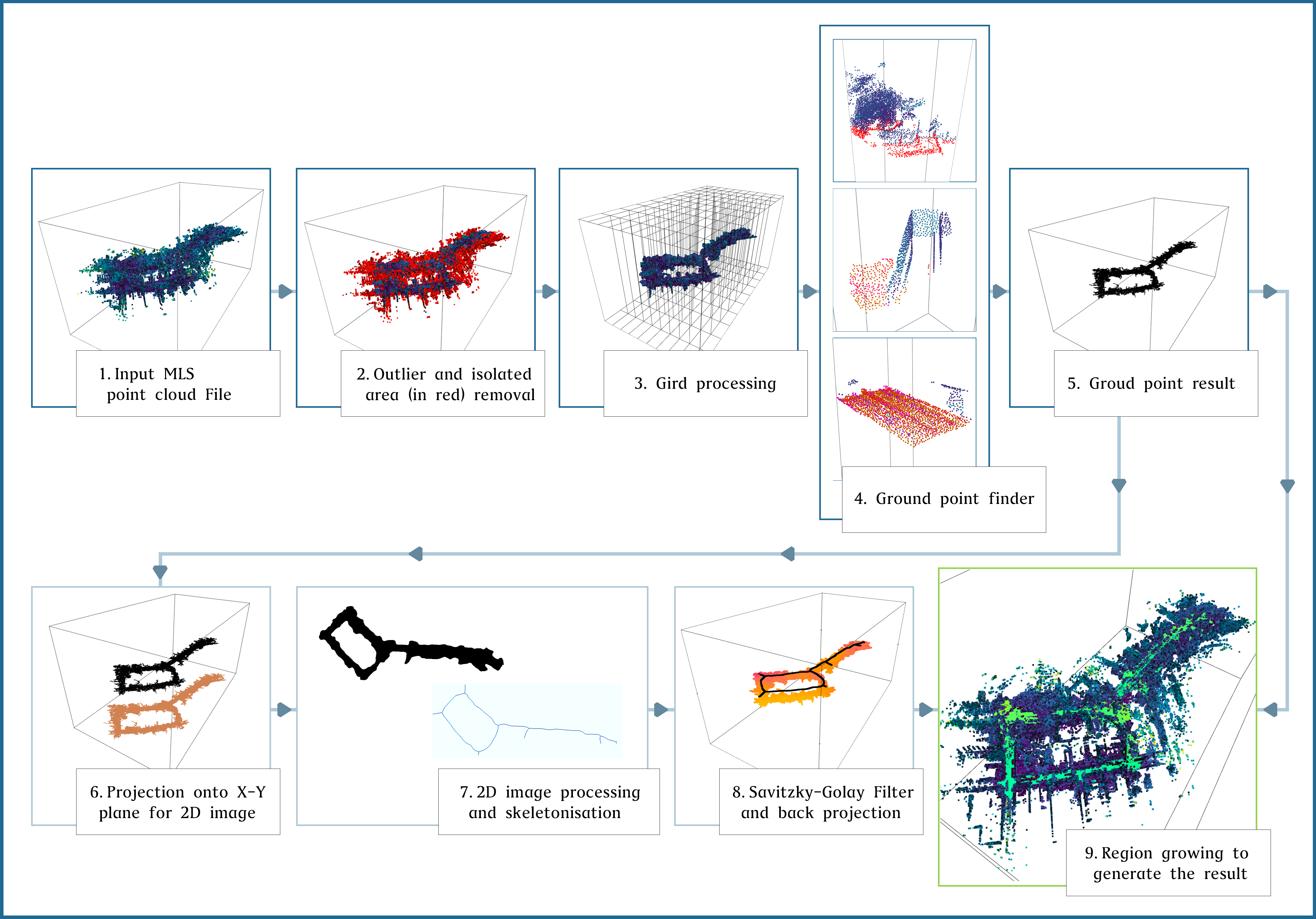}
    \caption{\small Workflow of point cloud data processing for feature extraction and analysis. (\textbf{1}) Initial input. (\textbf{2}) Removal of outliers and isolated areas to clean the data based on DBSCAN algorithm. (\textbf{3}) Grid processing that segments the point cloud into smaller parts. (\textbf{4}) Fitting of ground points using filtering methods. (\textbf{5}) Resultant ground points after processing. (\textbf{6}) Projection of the cleaned point cloud onto the XY-plane for 2D representation. (\textbf{7}) Further 2D image processing and skeletonisation to extract structural features. (\textbf{8}) Smoothing the framework by Savitzky-Golay filter followed by back projection to the original 3D space. (\textbf{9}) Normal-based region growing algorithm will be used on the filtered data to segment and generate the final processed result.}
    \label{pipeline}
\end{figure*}

\section {Proposed Approach} \label{Methodology}


 Fig.~\ref{pipeline} illustrates the pipeline of our proposed approach. First, DBSCAN and downsampling are applied to the point cloud data to eliminate noise and isolated areas. The point cloud is then segmented into smaller blocks on the XY plane. Within each block, the RANSAC algorithm is used to fit planes and assess their angles. Planes resembling road surfaces are retained based on angle criteria and majority coverage of the blocks, while other planes are filtered out using a moving threshold method, which preserves data from lower layers while discarding vertical planes. The road surface points collected from all blocks are further processed using DBSCAN to remove remaining isolated areas. Corner points are added to the extracted road point cloud to maintain alignment with the original raw data.

The processed data is then projected onto the XY plane to generate a top-down projection image, which is further refined through binarization and Gaussian blur to enhance structural visibility. Skeletonization is applied to extract centerlines, which are converted into scatter points and smoothed using the Savitzky-Golay filter \cite{Savitzky_golay} to form precise road skeletons. These 2D skeletons are re-projected onto the 3D point cloud, with calculated normals. Skeleton normals guide the region-growing algorithm to locate nearby road points. The proposed approach is tested on the Perth CBD point cloud dataset \cite{9647060}.

\begin{figure}
    \vspace{-20pt}
    \centering
    \includegraphics[width=1\linewidth]{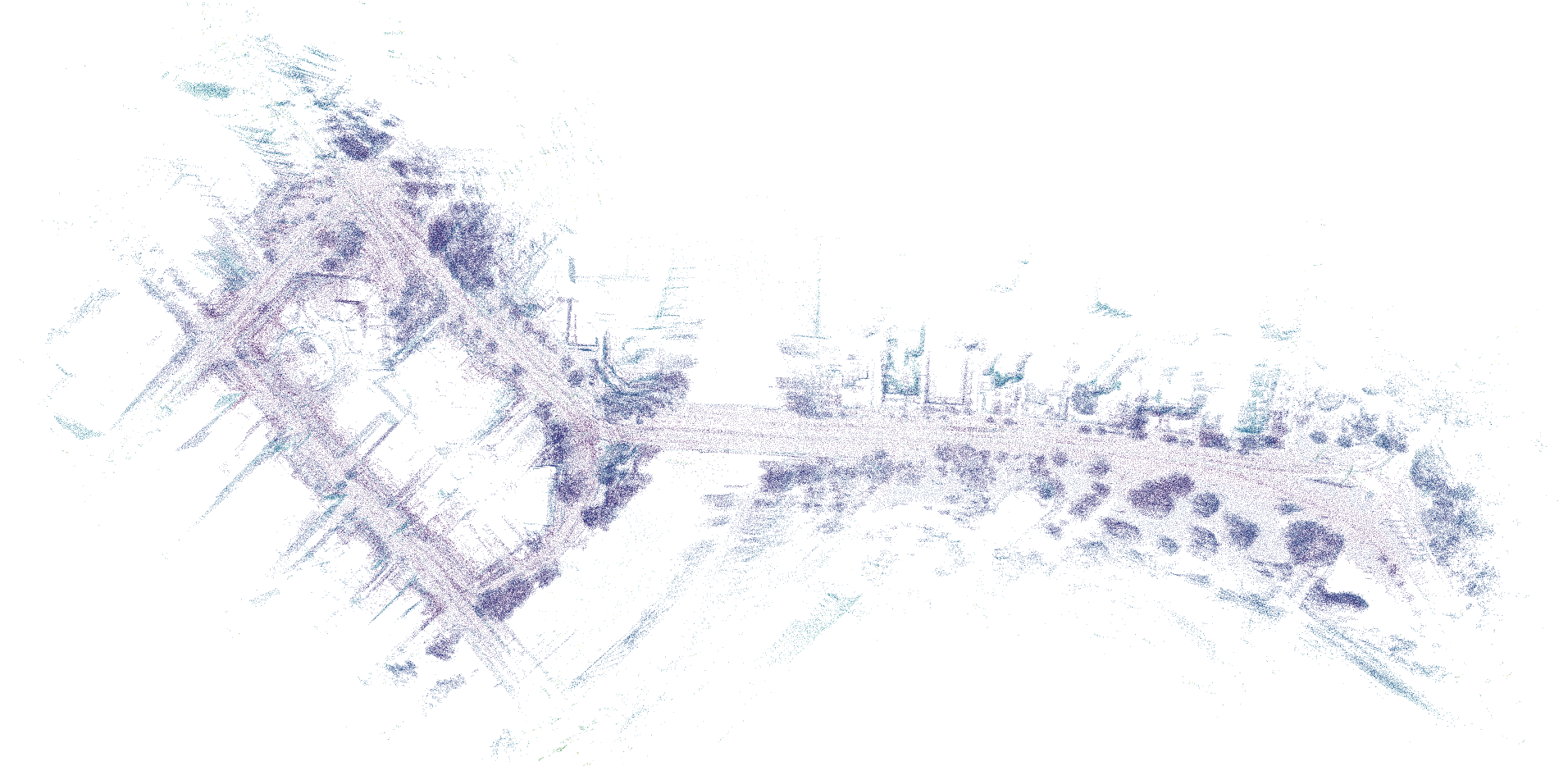}
    \caption{Example of an input MLS point cloud, containing road surfaces, vertical building facades and vegetation.}
    \label{fig-original}
    \vspace{-3mm}
\end{figure}

\begin{figure}
    \vspace{-20pt}
    \centering
    \includegraphics[width=1\linewidth]{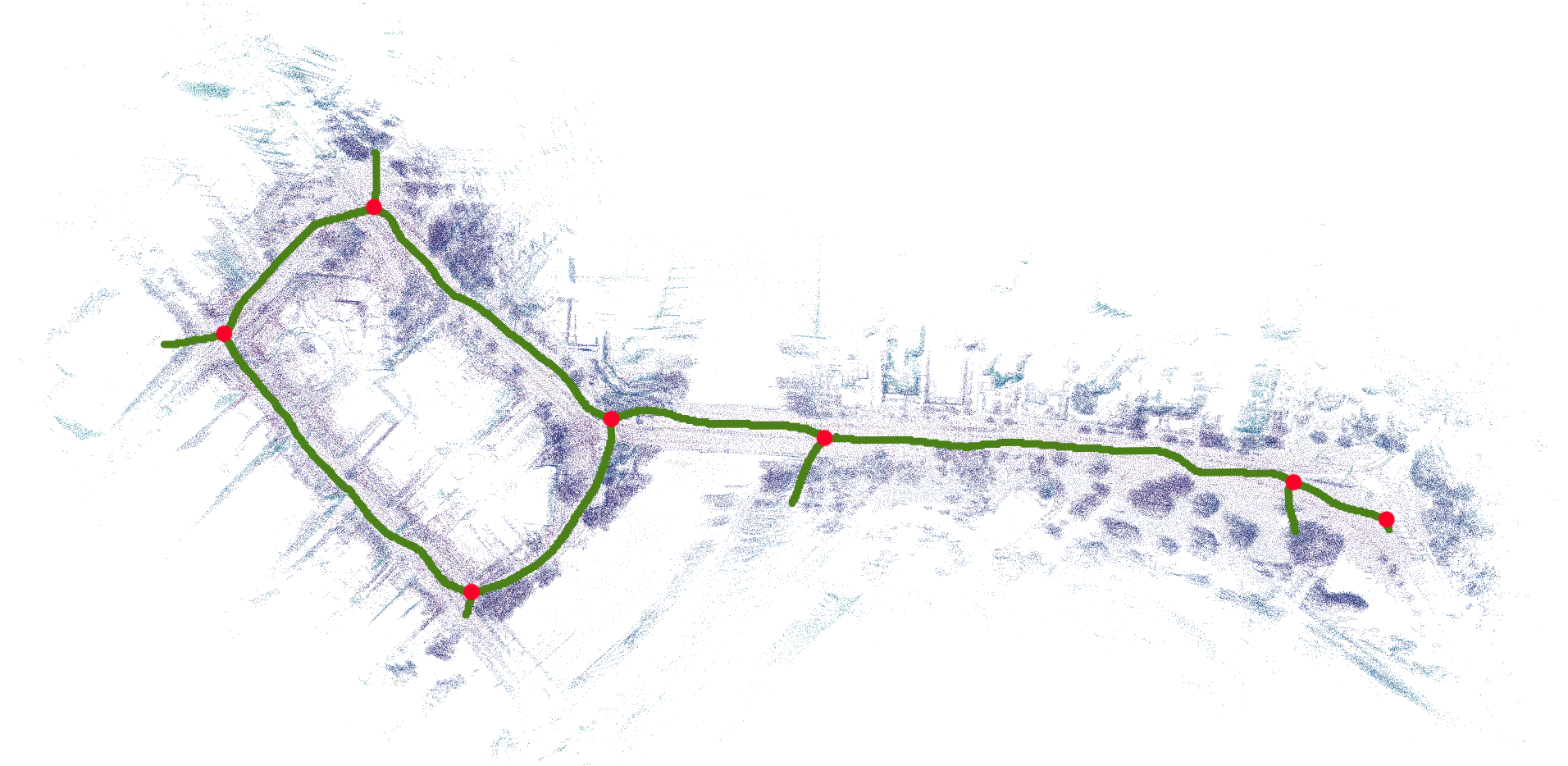}
    \caption{Skeletonisation result along with the original point cloud as the background for reference. Red dots indicates potential intersections}
    \label{skeleton_sample}
    \vspace{-3mm}
\end{figure}

\subsection{Preprocessing}

The first step in the proposed pipeline is to preprocess the input 3D point cloud data. The input data contains road surfaces, along with trees, buildings and other structures. Fig.~\ref{fig-original} provides an example of a MLS point cloud input file, which includes vertical building facades, road surfaces, vegetation, and low bushes. The primary goal of the preprocessing stage is to remove outliers and isolated areas, as they can negatively impact computational efficiency and lead to misleading estimations of the input data.

To address these issues, a two-step approach is employed. First, a distance-based statistical filtering method is used to remove noise points and outliers. This method relies on the distribution of distances between neighbouring points to identify and remove outliers. Mathematically, for each point $p_i$ in the point cloud, the mean distance $\mu_i$ to its k-nearest neighbours is calculated. Then, the overall mean $\mu$ and standard deviation $\sigma$ of all $\mu_i$ are computed. Points with $\mu_i$ outside the interval $[\mu - \alpha \sigma, \mu + \alpha \sigma]$, where $\alpha$ is a predefined parameter set at 2, are considered outliers and removed.

DBSCAN is applied to remove isolated small regions. DBSCAN groups together points that are closely packed, marking points in low-density areas as outliers \cite{10.5555/3001460.3001507}. It requires two parameters: $\epsilon$, the maximum distance between two points for them to be considered neighbours, and $minPts$, the minimum number of points required to form a dense region. Points that do not belong to any dense region are classified as isolated clusters and removed. We empirically set $\epsilon$ and $minPts$ to 2 and 10, respectively, to achieve a balanced outcome.

By applying these two preprocessing steps, the main road structure with attached areas is effectively maintained while noise and interference from outliers and isolated areas are significantly reduced as shown in step 2 of Fig.~\ref{pipeline}.

\subsection{Ground point filtering}

Roads can be non-planar in real data. Hence, our algorithm does not make any assumption of planar roads in the input data and can handle point clouds containing roads with varying slopes. 
To cater for elevation changes, we employ a grid-based segmentation, followed by processing of the small grid blocks.
This is depicted in step 3 of Fig.~\ref{pipeline}, where the entire point cloud is divided into small square chunks based on the X and Y dimensions, while preserving the original Z-axis coordinates of the points within each chunk.

The grid-based segmentation is performed by calculating the boundaries of each chunk using the following equations:

\begin{equation}
x_{min} = min(x_i), \quad x_{max} = max(x_i)
\end{equation}
\begin{equation}
y_{min} = min(y_i), \quad y_{max} = max(y_i)
\end{equation}
where $x_i$ and $y_i$ represent the X and Y coordinates of the points within the chunk, respectively. Each chunk is then processed individually for plane fitting using the RANSAC algorithm. RANSAC estimates the plane parameters by iteratively selecting random subsets of points and fitting a plane to them \cite{10.1145/358669.358692}. The plane with the highest number of inliers points within a specified distance threshold is considered the best-fitting plane. We empirically chose the threshold to be 30.




After estimating a flat surface for each chunk, the algorithm determines if the surface is part of the ground by comparing its tilt angle to the vertical direction. Chunks with angles exceeding 60\degree are classified as vertical structures (i.e., non-road points) and are discarded immediately. For chunks with smaller angles, the algorithm dynamically sets a filtering threshold based on the Z-value percentiles of the points within the chunk. Based on our experiments, we encountered three main types of grid block scenarios: the first type is when the whole block contains only road surface forming a perfect plane, the second type is a mixed situation with a road surface and vertical building structures, and the last one is a block containing unordered points. These three different situations are shown in step 4 of Fig.~\ref{pipeline}. This adaptive thresholding approach enables the algorithm to handle different road scenarios and terrain characteristics effectively, as depicted in step 5 of Fig.~\ref{pipeline}.

%



\subsection{Road points extraction}
\begin{figure}
    \vspace{-3mm}
    
    \centering
    \includegraphics[width=1\linewidth]{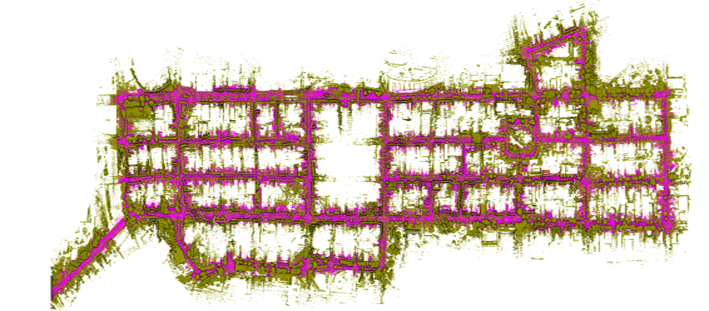}
    \caption{Preview of the Perth CBD point cloud data \cite{9647060}}
    \label{Perth_data_raw}
    \vspace{-3mm}
    
\end{figure}

\begin{figure*}
    \vspace{-25pt}
    \begin{subfigure}[b]{0.2\textwidth}
        \centering
        \includegraphics[width=\textwidth]{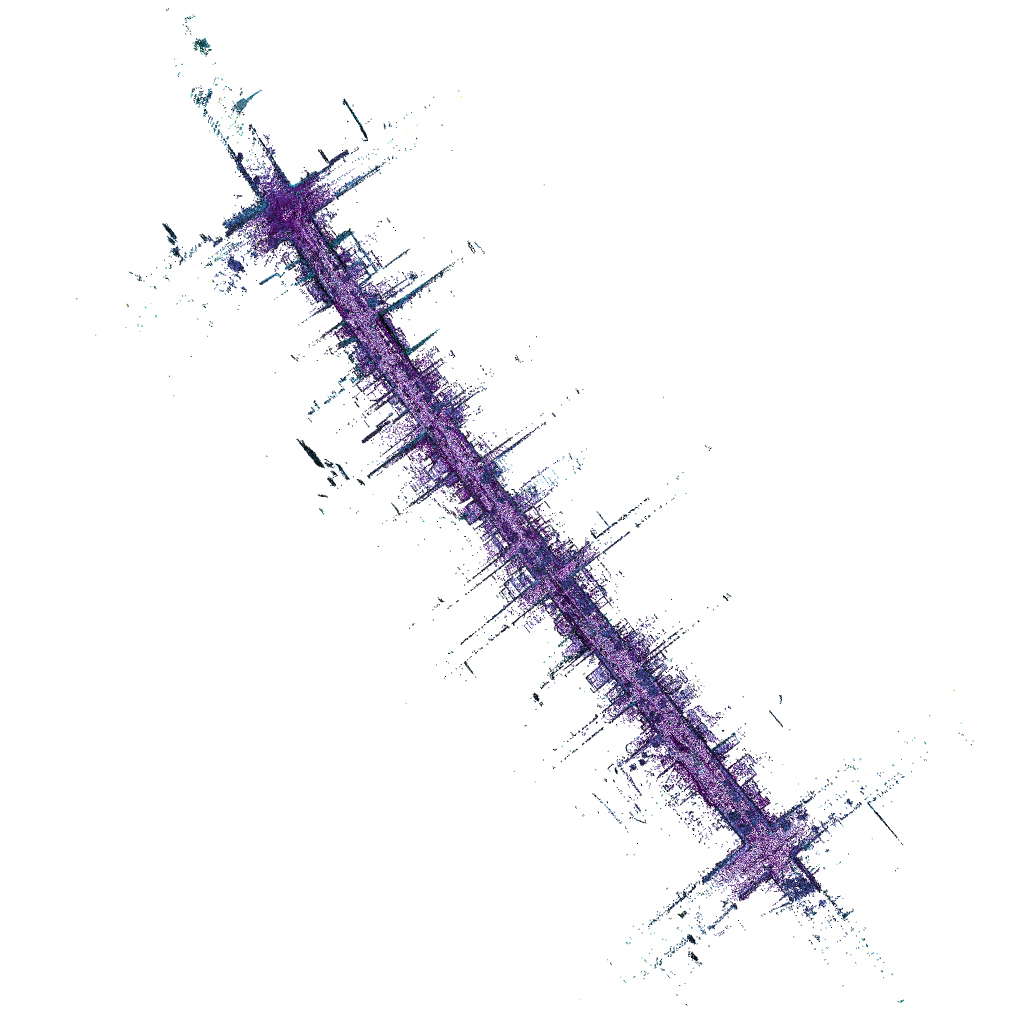}
        \caption{Selected file 10 is a straight road with driveways}
        \label{loop_demo1}
    \end{subfigure}
    \hfill
    \begin{subfigure}[b]{0.2\textwidth} 
        \centering
        \includegraphics[width=\textwidth]{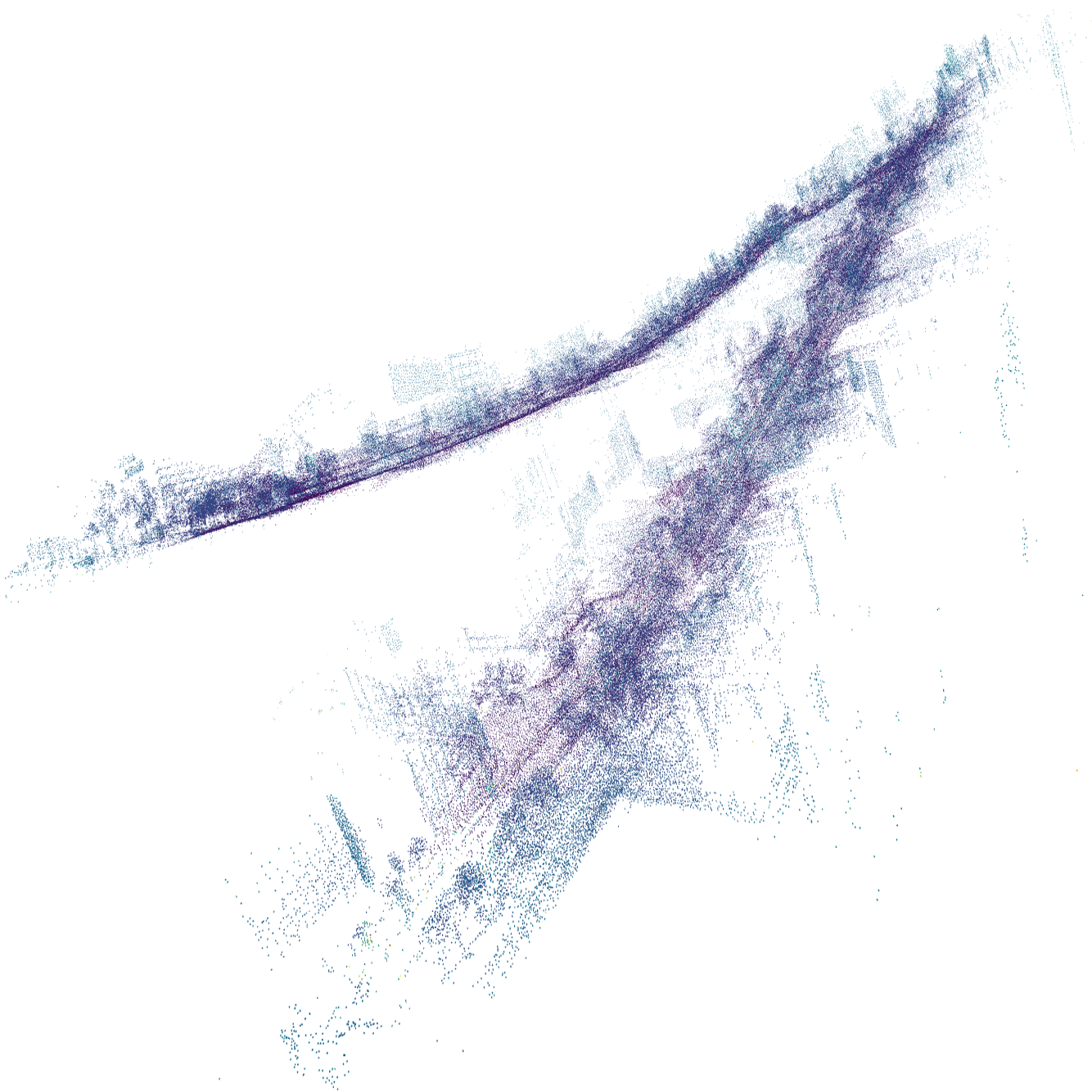}
        \caption{Selected file 9 contains accumulated errors}
        \label{loop_demo2}
    \end{subfigure}
    \hfill
    \begin{subfigure}[b]{0.2\textwidth}
        \centering
        \includegraphics[width=\textwidth]{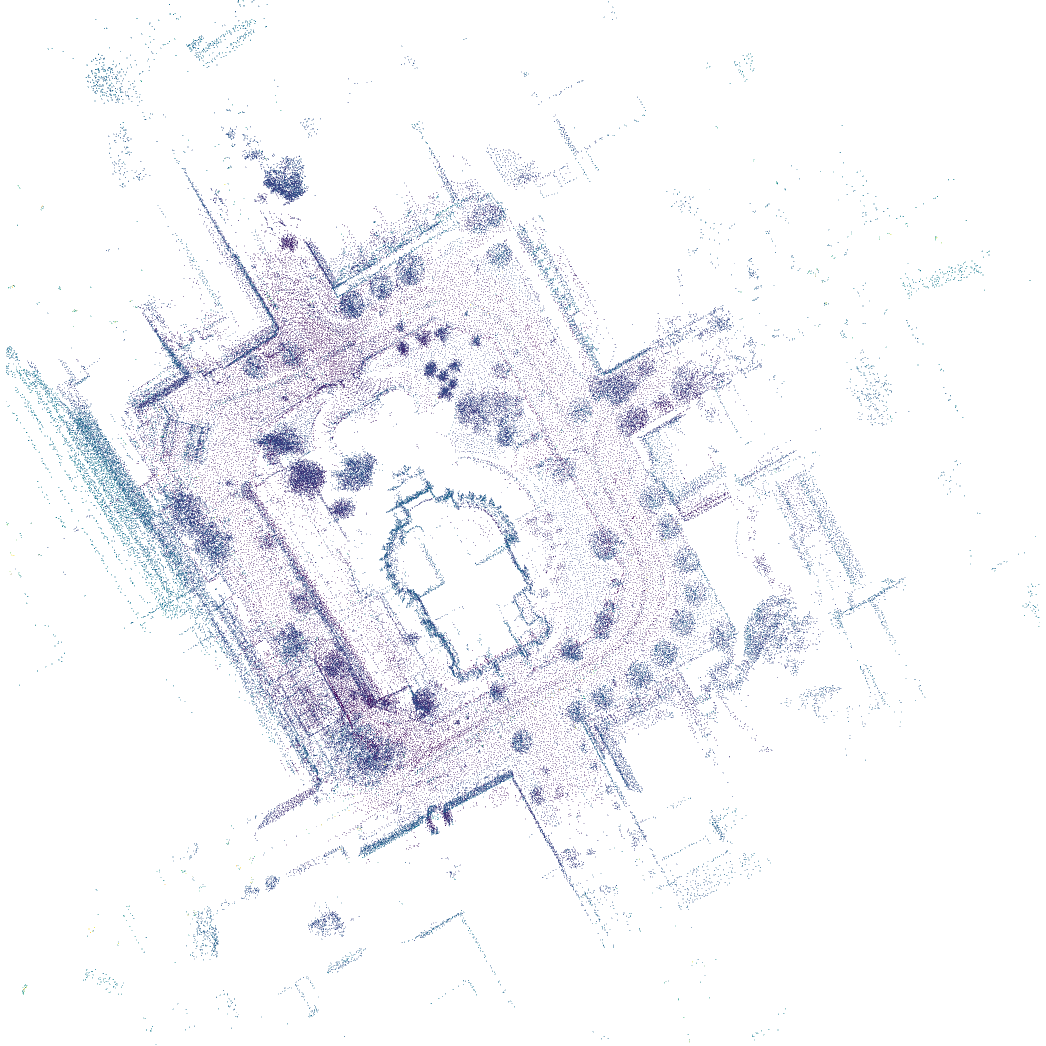}
        \caption{Selected file 11 is a ring road}
        \label{loop_demo3}
    \end{subfigure}
    \hfill
    \begin{subfigure}[b]{0.2\textwidth}
        \centering
        \includegraphics[width=\textwidth]{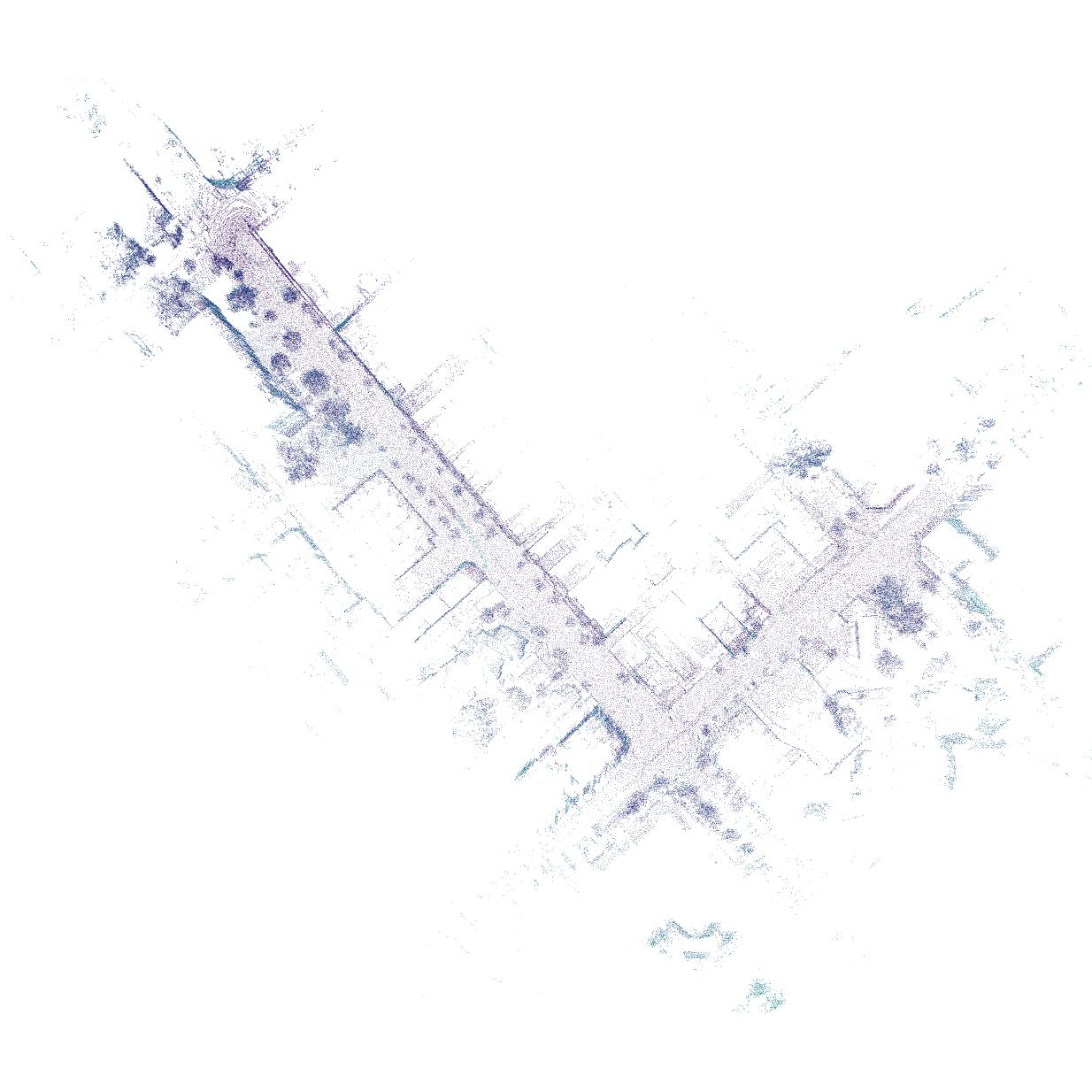}
        \caption{Selected file 13 has intersection}
        \label{loop_demo4}
    \end{subfigure}

    \caption{Previews for selected data with different road shapes and curvatures}
    \label{loop_demo}
    \vspace{-1pt}
\end{figure*}

The extraction of road points from the processed point cloud is performed by analysing the ground points from the top-down perspective. As shown in Fig.~\ref{Perth_data_raw}, the top-down view provides a clear overview of the road structure and facilitates the extraction of the road skeleton. By projecting the filtered point cloud onto the XY plane, the top-down perspective helps mitigate the skewness of the data, which is often caused by the urban canyon effect \cite{hough-1}. Drift and accumulated error due to the lack of GPS signal reception in dense urban areas contribute to the deformation of the final result \cite{9647060}. A sample of this type of data is shown in Fig.~\ref{loop_demo2}.

However, the projected image may contain significant noise that needs to be removed. To tackle this issue, Gaussian filtering is employed for denoising. The Gaussian filter performs smoothing by convolving the data with a Gaussian kernel, defined as:
\begin{equation}
    G(x, y) = \frac{1}{2\pi\sigma^2} e^{-\frac{x^2 + y^2}{2\sigma^2}}
\end{equation}
where $\sigma$ is the standard deviation of the Gaussian distribution. The choice of $\sigma$ determines the level of smoothing applied to the image. In this study, the $\sigma$ value is preset at 5 to achieve a balance between noise reduction and preserving essential road structure details.

After denoising, the skeleton of the road structure is extracted from the binary image. The skeletonisation process is performed using OpenCV based iterative thinning approach \cite{opencvpython2012skeletonization}. The algorithm iteratively erodes the boundary pixels of the road structure while preserving its connectivity. The result is shown in Fig.~\ref{skeleton_sample} with the original point cloud file as the background for reference.

The skeleton is back-projected onto the 3D point cloud with moving Z-values, allowing it to align perfectly with the original road section, even in cases of high curvature. Normals are then calculated based on the skeleton, now converted to point cloud format. A region-growing algorithm using the K-D Tree method is subsequently applied to identify road points that are perpendicular to the skeleton normals and share similar Z-values. These extracted road points are then skeletonised again to generate the road centerline, as shown in subfigure (d) of Figs.~\ref{loop_process_9} - \ref{loop_process_13}. This approach extracts the road points and road centreline from the top-down perspective, corrects data skewness and generates a smooth road centreline. 


\section {Experiment and results} \label{Experiment}
The proposed algorithm is tested on the Perth CBD point cloud dataset \cite{9647060}. The dataset covers approximately 4 square kilometres of the Perth city centre, containing over 64 million 3D points, making it a large-scale, high-precision urban point cloud dataset. In the complete 3D map shown in Fig.~ \ref{Perth_data_raw}, roads are annotated in pink, while other objects such as buildings, traffic lights, trees, and lamp posts are annotated in green, providing rich semantic information. 

\begin{figure}
    \centering
    \includegraphics[width=1\linewidth]{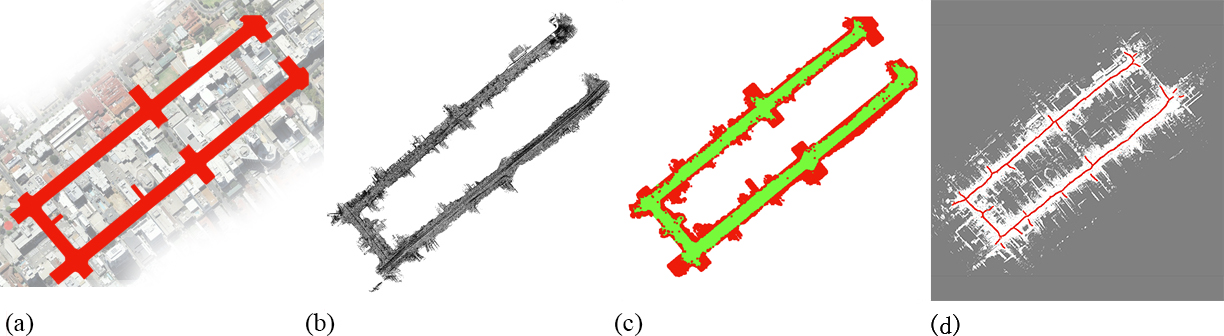}
    \caption{Sample process result for input file 9 (\textbf{a}) Ground truth with satellite image (\textbf{b}) Extracted road points (\textbf{c}) IoU result visualisation (\textbf{d}) Centreline fitting result. }
    \label{loop_process_9}
    \vspace{-5mm}
\end{figure}

\begin{figure}
    \centering
    \includegraphics[width=1\linewidth]{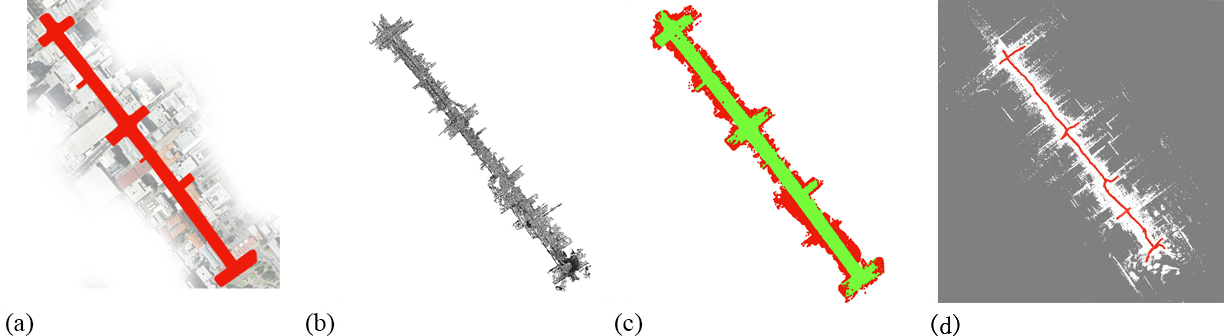}
    \caption{Sample process result for input file 10 (\textbf{a}) Ground truth with satellite image (\textbf{b}) Extracted road points (\textbf{c}) IoU result visualisation (\textbf{d}) Centreline fitting result. }
    \label{loop_process_10}
    \vspace{-5mm}
\end{figure}

\begin{figure}
    \centering
    \includegraphics[width=1\linewidth]{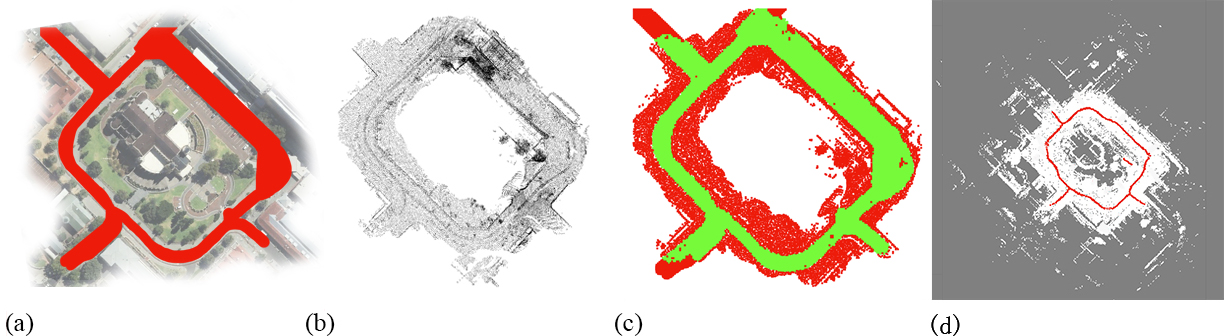}
    \caption{Sample process result for input file 11 (\textbf{a}) Ground truth with satellite image (\textbf{b}) Extracted road points (\textbf{c}) IoU result visualisation (\textbf{d}) Centreline fitting result. }
    \label{loop_process_11}
    \vspace{-5mm}
    
\end{figure}

\begin{figure}
    \centering
    \includegraphics[width=1\linewidth]{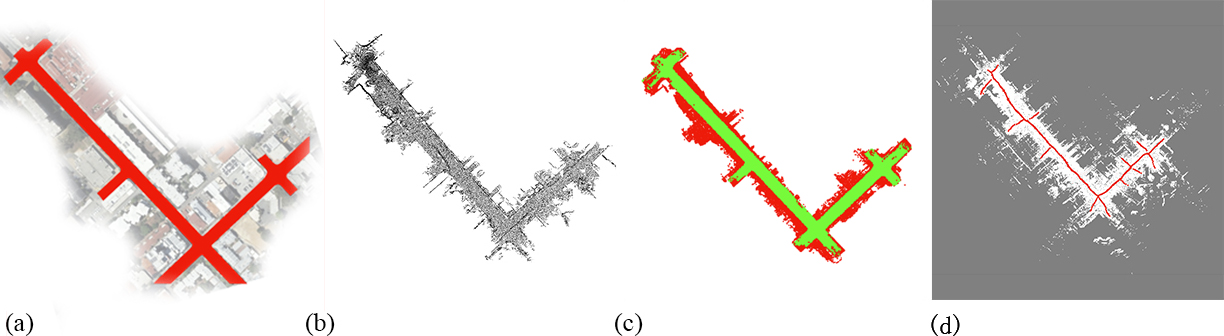}
    \caption{Sample process result for input file 13 (\textbf{a}) Ground truth with satellite image (\textbf{b}) Extracted road points (\textbf{c}) IoU result visualisation (\textbf{d}) Centreline fitting result. }
    \label{loop_process_13}
    \vspace{-5mm}
\end{figure}

Although the Perth CBD dataset provides registered and large-scale point clouds in a single file, we used smaller registered files to conduct our experiments. These are individual point cloud files containing reduced number of points, covering different areas of the Perth CBD. The dataset has different smaller size files with varying road structures, and demonstrated effectiveness of our approach as shown in subfigures (b) and (d) of the Figs.~\ref{loop_process_9} - ~\ref{loop_process_13}. 

\begin{figure}[H]
    \includegraphics[width=1\linewidth]{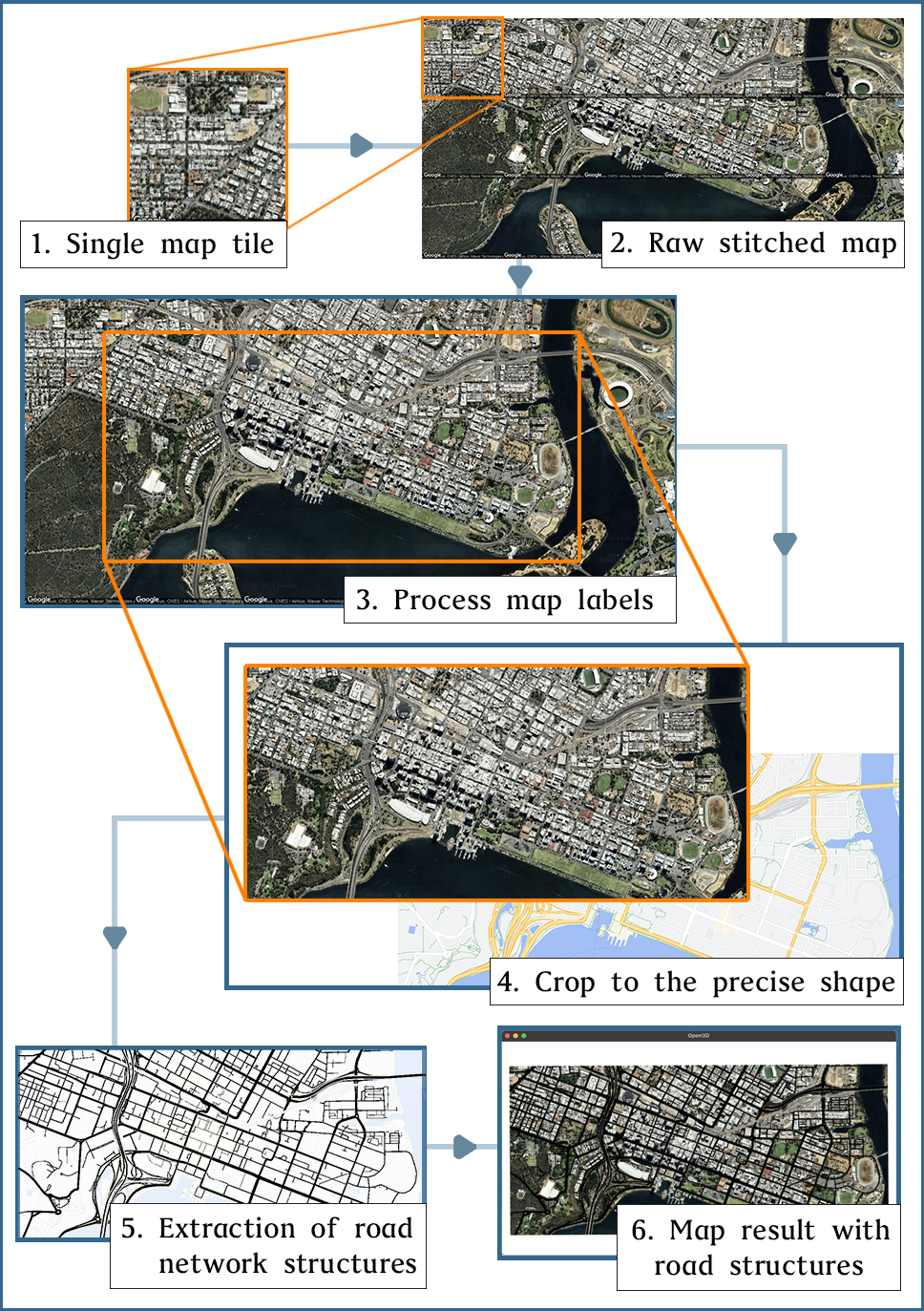}
    \caption{\small Workflow of obtaining ground truth from GoogleMaps (\textbf{1}) Identified batch of single map tiles based on the given coordinates and the zoom level (\textbf{2}) Original stitched map created by calculating coordinates for individual tiles based on the tile size  and the ratio of given GPS coordinates boundaries (\textbf{3}) Remove labels from map tiles, only labels affecting the road structures are removed. (\textbf{4}) Crop based on the pixel/coordinate ratio This is the precise image we want for the  given coordinates (same process for the map) (\textbf{5}) Extraction of road network structures by  implementing colour filtering and skeletonisation processing from openCV library (\textbf{6}) The ground truth is created with precise coordinates}
    \label{mappipeline}
    \vspace{-3mm}
\end{figure}

We used data from Google Maps as the ground truth source. The original Google API provides a maximum resolution of 640x640 pixels for map tiles, which is insufficient for our evaluation. To address this, we developed an automated method to obtain satellite images and 2D road structures from Google Maps for any given coordinate boundary at any zoom level by stitching map tiles together. Figure \ref{mappipeline} illustrates the process for generating the stitched satellite image and its extracted road network layer from the corresponding map. First, a batch of individual map tiles is identified based on the given coordinates and zoom level. Then, a complete map is created by stitching these tiles together using the tile size and GPS coordinate boundaries. Each tile includes a watermark located at the bottom. To avoid interference, labels on the bottom of the tiles are removed by adjusting the coordinates and shifting subsequent rows of tiles upward by 22 pixels. Only labels that interfere with the road structure extraction process are removed; labels on the last row of tiles are always retained. The map is then cropped according to the pixel-to-coordinate ratio to achieve the desired image for the specified coordinates. Finally, road network structures are extracted using color filtering and skeletonization processes from the OpenCV library, producing accurate ground truth road structure masks.

The mask is used to calculate Intersection over Union (IoU) values and to facilitate the visual comparison. The calculation of the IoU value is mainly based on the K-D Tree approach proposed by Skrodzki et. al. \cite{kdtree} to find the nearby points based on the downsampled result. 
The IoU calculation algorithm is detailed in Algorithm \ref{al_iou}.

\begin{algorithm}
\caption{Calculation of IoU}
\begin{algorithmic}[1]

\State \textbf{Input:} cloud1, cloud2, threshold
\State \textbf{Output:} IoU value

\Procedure{CalculateIoU}{$cloud1, cloud2, threshold$}

\State Voxel down-sample $cloud1$ and $cloud2$
\State Convert down-sampled clouds to point arrays

\State Build KD-Trees for both point arrays

\State Find intersection by searching nearby points in KD-Trees

\State Calculate union by merging two point sets

\State IoU $\gets$ Size of intersection / Size of union

\State \textbf{return} IoU

\EndProcedure
\end{algorithmic}
\label{al_iou}
\end{algorithm}

For demonstration purpose, we selected four representative scenes from 17 pre-merged point cloud files: road intersections, looped road, straight road, and data with large cumulative errors. Fig.~\ref{loop_demo} shows previews of these four scenes, illustrating their distinct road shapes and characteristics. 
By adopting the skeletonisation function from scikit-image library \cite{sk}, key nodes within the road network, such as intersections, were correctly identified. This was achieved using a 3x3 convolution kernel to detect cross-points in the framework image generated by projection, facilitating an understanding of road layouts and revealing significant branching points within the network.


\begin{table*}
    
    \centering
    \caption{Comparison of performance of post-skeleton process}
    \label{tab:comparison_data}
    \setlength{\tabcolsep}{4pt} 
    \begin{tabular}{@{}lrrrrrrr@{}}
        \toprule
            File Name & Original Points & Previous Points & Current Points & Point Reduction (\%) & Previous Time (s) & Current Time (s) & Time Improvement (\%) \\ 
        \midrule
            Loop9.ply & 392264 & 284358 & 113514 & 60.08 & 13.32 & 15.36 & -15.32 \\
            Loop10.ply & 143711 & 97651 & 40815 & 71.60 & 3.61 & 2.16 & 40.17 \\
            Loop11.ply & 206827 & 165327 & 69082 & 58.21 & 6.87 & 4.44 & 35.37 \\
            Loop13.ply & 204134 & 151242 & 59615 & 60.57 & 6.21 & 4.21 & 32.21 \\
        \midrule
            Ave. Improvement & - & - & - & 62.62 & - & - & 23.11 \\
        \bottomrule
        
    \end{tabular}
\end{table*}

During the experiments, we tested the algorithm's performance without the refinement of connecting endpoints in the generated skeletons. As the original generated skeleton graph contains disconnected branches, we calculated endpoint distances of each separated branch and connect points together based upon their Euclidean distances. Once the skeleton is processed, disconnected small branches were discarded.



These enhancements lead to a substantial reduction in the number of processed points which is at least 60\% less than the previous approach, and also decreased the processing time significantly by at least 23\%. The comparative data presented in Table \ref{tab:comparison_data} showing our approach's improvement in processing efficiency. Overall, these changes boost the generalisation capabilities of our approach across different types of input data.


We implemented a grid-based segmentation approach, combining RANSAC plane fitting with moving median filtering to extract ground point clouds. The ground points are extracted comprehensively, and the moving threshold algorithm effectively removes non-ground planes with large angles while preserving smooth and continuous road surface point clouds. Even in cases of severe deformation caused by cumulative errors, the filtered ground point cloud shown in subfigure (b) of Figs. \ref{loop_process_9} - \ref{loop_process_13} maintained good ground details. Compared to the original point cloud in Fig. \ref{loop_demo}, ground point filtering not only retains the road structure but also removes irrelevant points with reduced dimension size.

We then project the processed point cloud onto the XY plane for skeleton extraction and centreline fitting. The calculated nromals from the skeleton guides the region-growing algorithm to pick up road points. The result is shown in subfigure (b) of Figs.~\ref{loop_process_9} - \ref{loop_process_13}. To further smooth the skeleton, we apply the Savitzky-Golay filter for local polynomial fitting of the skeleton points, obtaining the final road centreline at this stage in subfigure (d) of Figs.~\ref{loop_process_9} - \ref{loop_process_13}.


Table~\ref{table_result} presents IoU results for various road scenarios, comparing previous and current approaches. Loop files 9 and 10 exhibit higher IoU scores of 82.5\% and 81.5\%, respectively, attributed to fewer open areas surrounding the roads in these scenarios. This improvement in IoU was facilitated by post-processing the extracted skeletal framework. By trimming and connecting the skeleton and calculating normals, the region-growing algorithm more effectively identified neighboring points close to the skeleton with similar Z-values, resulting in a clearer delineation of road points.

In loop file 11, as shown in Fig.~\ref{loop_process_11}c, areas painted in red depict erroneously recognized road surfaces due to two main reasons. First, the presence of extensive green spaces in the image shares similar colors and textures with the road, making it challenging for the algorithm to differentiate between them. Second, the modified one-way road is connected to these green spaces, further complicating the accurate delineation of boundaries between road and non-road areas. The current approach improved the IoU result from 54.8\% to 61.2\%, the largest improvement across these scenarios. However, it still presents challenges in distinguishing the boundary between open areas and the road.

Overall, the extracted road point clouds are generally wider than the ground truth masks. The primary cause of this phenomenon is likely the low resolution of the point clouds, which hinders the algorithm from successfully identifying the kerb, i.e., the road edge. In the absence of distinct boundary features, open areas adjacent to the road, such as car parks, pedestrians, and low-lying bushes, are tend to misclassification as road points.

\begin{table}[htbp]
\caption{Comparison of IoU Metrics for Selected Point Cloud Data}
\label{table_comparison}
\centering
\begin{tabular}{@{}lccc@{}}
\toprule
File Name & Previous IoU & Current IoU & Improvement (\%) \\ 
\midrule
Loop\_9.ply  & 0.766 & 0.825 & \textbf{7.70} \\
Loop\_10.ply & 0.748 & 0.815 & \textbf{8.96} \\
Loop\_11.ply & 0.548 & 0.612 & \textbf{11.68} \\
Loop\_13.ply & 0.653 & 0.698 & \textbf{6.89} \\
\bottomrule
\end{tabular}
\label{table_result}
\end{table}

The results shown that we evaluated each component of the proposed algorithm through experiments. The experimental results in Table~\ref{table_result} proved the effectiveness of improvement, this will be the foundation for further refinement of road point extraction result and more in-depth road analysis and modelling.

\section { conclusion and Way forward } \label{conclusion}

The proposed road extraction algorithm has demonstrated promising results in the experiments, indicating its capability to accurately extract road centrelines and relevant points from extensive urban point cloud datasets using a combined 3D-2D pipeline and the top-down strategy. Enhancements, such as post-skeleton processing and dimension management, have improved the approach, better preserving road points near the centrelines. We utilise normals to determine road orientation and employ vectors orthogonal to these normals to enhance the region-growing algorithm in our approach. The IoU metric for the latest approach shows an average improvement of 8.8\% using four different road scenarios from the Perth CBD dataset. Moreover, the processing time has decreased by an average of 23\%. Our approach performs well to various road shapes and scenarios, demonstrating its versatility and effectiveness.


However, the algorithm still has some limitations that require further improvements and optimisations. The algorithm is currently sensitive to parameter settings, with the choice of grid size and RANSAC thresholds significantly impacting the results. The extracted road framework shown in 2D format still contains some outliers, affecting the accuracy of the results. 

Future research directions include separating open areas like car parks from extracted road points, optimising RANSAC parameters to adapt to different road scenarios, refining region-growing algorithm to detect street directions and pedestrian pathway boundaries for better results. 
Despite limitations, the algorithm holds promise for applications in city mapping and traffic planning.

\section*{acknowledgments}Professor Ajmal Mian is the recipient of an Australian Research Council Future Fellowship Award (project number FT210100268) funded by the Australian Government.

\bibliographystyle{IEEEtran}
\bibliography{IEEEabrv,frank_dicta}

\end{document}